\documentclass[lettersize,journal]{IEEEtran}
\usepackage{amsmath,amsfonts}
\usepackage{algorithmic}
\usepackage{algorithm}
\usepackage{array}
\usepackage[caption=false,font=normalsize,labelfont=sf,textfont=sf]{subfig}
\usepackage{textcomp}
\usepackage{stfloats}
\usepackage{url}
\usepackage{verbatim}
\usepackage{graphicx}
\usepackage{float}
\usepackage{booktabs}
\usepackage{multirow}
\usepackage{cite}

\usepackage{multirow}
\begin{document}

\title{Resolution Enhancement of Under-sampled Photoacoustic Microscopy Images using Implicit Neural Representations}

% \author{Youshen Xiao,~\IEEEmembership{Student member,~IEEE,} 
%         Yuyao Zhang*,~\IEEEmembership{Member,~IEEE,}
%         Fei Gao*,~\IEEEmembership{Member,~IEEE,}

\author{Youshen Xiao,
        Sheng Liao,
        Xuanyang Tian,
        Fan Zhang,
        Xinlong Dong,
        Yunhui Jiang,
        Xiyu Chen,
        Ruixi Sun,
              
        Yuyao Zhang*,
        Fei Gao*
        % <-this % stops a space
\thanks{Youshen Xiao, Shen Liao, Xuanyang Tian, Fan Zhang, Xinlong Dong, Yunhui Jiang, Xiyu Chen, Ruixi Sun are with the School of Information Science and Technology, ShanghaiTech University, Shanghai 201210, China (e-mail: xiaoysh2023@shanghaitech.edu.cn; ).}% <-this % stops a space
\thanks{Yuyao Zhang are with the School of Information Science and Technology and Shanghai Engineering Research Center of Intelligent Vision and Imaging, ShanghaiTech University, Shanghai 201210, China (email: zhangyy8@shanghaitech.edu.cn).}
\thanks{Fei Gao was with the School of Information Science and Technology, ShanghaiTech University, Shanghai, 201210, China. He is now with the School of Biomedical Engineering, Division of Life Sciences and Medicine, University of Science and Technology of China, Hefei, Anhui, 230026, China, and with the Suzhou Institute for Advanced Research, University of Science and Technology of China, Suzhou, Jiangsu, 215123, China (e-mail: xjtugaofei@foxmail.com).

%; liaosheng2022@shanghaitech.edu.cn; tianxy@shanghaitech.edu.cn; zhangfan2022@shanghaitech.edu.cn; jiangyh2022@shanghaitech.edu.cn; sunrx2023@shanghaitech.edu.cn

}% <-this % stops a space
}

\markboth{Journal of \LaTeX\ Class Files,~Vol.~14, No.~8, August~2021}%
{Shell \MakeLowercase{\textit{et al.}}: A Sample Article Using IEEEtran.cls for IEEE Journals}

\IEEEpubid{ }

\maketitle

\begin{abstract}
Acoustic-Resolution Photoacoustic Microscopy (AR- PAM) has shown potential in subcutaneous vascular imaging. However, the spatial resolution of its images is limited by the system’s Point Spread Function (PSF). To improve resolution, various deconvolution-based methods can be employed. Traditional deconvolution methods, such as Richardson-Lucy deconvolution and model-based deconvolution, use the PSF as prior knowledge to enhance spatial resolution. However, accurately measuring the system’s PSF is challenging, leading to the commonly used blind deconvolution methods, which often suffer from inaccurate deconvolution. Another challenge of AR-PAM is the long scanning time. To speed up the image acquisition, down-sampling can be used to reduce scanning time. Subsequently, interpolation methods are applied to recover the high-resolution image from down-sampled image; however, traditional interpolation methods are insufficient for high-fidelity image recovery, especially under high under-sampling conditions. In this work, we propose a method based on Implicit Neural Representations (INR) to address both of these issues: unknown PSF and under-sampled image recovery. By using INR, we learn a continuous mapping from spatial positions to initial acoustic pressure, effectively compensating for the discretization of the image space and enhancing the resolution of AR-PAM. Specifically, we set the PSF as learnable parameter to address the issue of inaccurate PSF measurement. We qualitatively and quantitatively evaluated the proposed method on simulated vascular data, demonstrating superior performance in metrics such as Peak Signal-to-Noise Ratio (PSNR) and Structural Similarity Index (SSIM) compared to traditional methods. Additionally, we showed qualitative improvements on leaf vein data and \textit{in vivo} mouse brain microvasculature.
\end{abstract}


\section{Introduction}
\IEEEPARstart{P}{hotoacoustic} Imaging (PAI) is a non-invasive biomedical imaging technique that combines the advantages of optical absorption contrast with the deep penetration capabilities of ultrasound \cite{1,2}. In PAI, acoustic waves are generated by the thermoelastic expansion caused by transient light absorption, which are subsequently captured by ultrasonic sensors to construct PA images, representing the original distribution of initial pressure\cite{3,4}. With multi-wavelength light illumination, PAI exhibits high sensitivity to optical absorption contrast. Theoretically, any changes in the optical absorption coefficient directly reflects in the intensity variations of the detected PA signals, forming the basis for its functional and molecular imaging capability, such as the measurement of hemoglobin oxygen saturation (sO2)\cite{5}. Depending on the imaging system configuration, PAI is primarily categorized into three forms: Photoacoustic Microscopy (PAM), Photoacoustic Computed Tomography (PACT), and Photoacoustic Endoscopy (PAE).

PAM offers high-resolution imaging within subcutaneous depth, showcasing significant optical absorption contrast and revealing structural, functional, and molecular information of the tissue\cite{6}. Optical-resolution PAM (OR-PAM), a variant of PAM, achieves near-diffraction-limited resolution through the use of tightly focused laser beams for excitation. Over the past decade, OR-PAM has garnered significant attention due to its superior imaging capabilities and has been widely applied in neuroscience research, tumor angiogenesis monitoring, histology examinations, dermatological diagnostics, and other preclinical and clinical studies\cite{7,8,9}. On the other hand, acoustic-resolution PAM (AR-PAM), leveraging the good penetration of scattered light and sound waves in deeper tissues, demonstrates greater advantages in deep-tissue imaging compared to OR-PAM\cite{10}, which has proven its value in microvascular imaging\cite{11}.

The lateral resolution of AR-PAM is determined by the center frequency and numerical aperture (NA) of the focused acoustic transducer used. To achieve higher lateral resolution, AR-PAM can employ transducers with higher central frequency and NA. However, this approach comes with challenges, such as requiring the system to detect high-frequency acoustic waves, which experience significant attenuation in biological tissues, thus affecting their ability to penetrate deeply. Additionally, a higher NA can also limit the depth of focus, making it more difficult to maintain good focus over a certain depth range. To enhance the lateral resolution of AR-PAM while overcoming the issues mentioned above, deconvolution algorithms can be employed.

In recent years deconvolution algorithms have been widely applied in the field of photoacoustic imaging, including PACT\cite{15}, OR-PAM\cite{12,13}, and AR-PAM\cite{14}. For AR-PAM, various deconvolution techniques have been proposed, notably including Richardson-Lucy (R-L) deconvolution and Malvar-Burton (MB) deconvolution. From the perspective of an image degradation model, restoring a clear image fundamentally involves a deconvolution process, which relies on the estimation of the Point Spread Function (PSF). In AR-PAM, the PSF is mainly determined by the characteristics of the acoustic transducer's focal region.

Another notable aspect of AR-PAM is that its performance is primarily constrained by two interrelated key factors: imaging speed and spatial resolution. According to the Nyquist sampling theorem, to ensure that PAM provides high spatial resolution images, the scanning step must be less than half of the lateral resolution to prevent aliasing, which would otherwise degrade image quality. Therefore, traditional PAM requires a large amount of sampling data, which increases acquisition and processing time, raises system memory requirements, and thus the cost. To achieve fast PAM systems without significantly reducing resolution or increasing costs, researchers have developed advanced data acquisition schemes and innovative signal processing methods\cite{16,17}. For AR-PAM, different scanning mechanisms can be adopted depending on the required imaging speed\cite{18}. For instance, when video-rate three-dimensional imaging is needed, a raster scan of the excitation laser beam can be performed within the acoustic focus (typically with a diameter of about 50 micrometers), although this limits the field of view. To further increase imaging speed, a digital micromirror device can be used for random-access scanning of specific features within the region of interest, effectively skipping background areas. Nevertheless, even with these approaches, data acquisition speeds are still limited by point-by-point optical scanning. Recently, hybrid scanning using water-immersed microelectromechanical systems mirrors has achieved three-dimensional imaging rates of approximately 1Hz with a moderate field of view (about 3x4 square millimeters) while maintaining high detection sensitivity. This method enhances both imaging speed and quality by simultaneously focusing the excitation laser beam and the received acoustic waves. At the algorithmic level, a sparse matrix recovery method based on the alternating direction method of multipliers has been proposed for sparse optical scanning PAM systems to enable rapid vascular imaging.

In recent years, deep learning methods have emerged in the field of photoacoustic imaging\cite{19}. By training neural network models, these methods can automatically extract features from raw data and generate high-quality images without the need for traditional signal processing steps. Deep learning has shown remarkable performance in addressing issues such as sparse-view, limited-view, and artifact removal in PACT; it has also been applied to deconvolution and undersampling problems in PAM. Chen et al.\cite{20} proposed using deep learning to transform blurred \textit{in vivo} mouse vasculature images acquired by AR-PAM systems to achieve deep-penetration OR-PAM performance. Feng et al.\cite{21} aiming to enhance the lateral resolution of AR-PAM, developed a multi-scale feature high-fidelity restoration algorithm based on deep convolutional neural networks. To accelerate PAM imaging, M. Burcin Unlu et al. introduced a novel and flexible algorithm called DiffPam\cite{16}, based on diffusion models, to speed up the photoacoustic imaging process. Although deep learning methods are effective, improving the performance of AR-PAM deconvolution algorithms or accelerating PAM imaging speeds requires substantial training data. Given the relatively high cost of acquiring PAM data and the difficulty in obtaining paired datasets, there remains a need for more efficient and practical algorithms to further enhance PAM imaging speed and resolution.

Implicit Neural Representations (INR) represent a new approach that parameterizes signals using Multi-Layer Perceptron (MLP)\cite{22}. Unlike traditional explicit representations that use discrete elements such as pixels or voxels, INR represent the object itself as a continuous function of spatial coordinates. That is, the values at any spatial positions of the object can be retrieved by querying the corresponding coordinates of a trained MLP. This provides a universal solution for various applications in object reconstruction. With the application of MLP and appropriate encoding functions that map input coordinates into high-dimensional spaces\cite{23}, INR have achieved superior performance in multiple computer vision tasks\cite{23,24}. Previous studies have also shown that INR can solve inverse problems in the medical imaging domain in an unsupervised manner, such as CT image reconstruction\cite{25,26} and undersampled MRI\cite{27}. Our previous work utilized INR to reconstruct photoacoustic images under sparse and limited-view conditions\cite{28}.

In this work, we explored the use of INR to accelerate the image reconstruction process in PAM and improve the resolution of AR-PAM. The effectiveness of INR-based sparse sampling deconvolution was first validated through simulated vascular images. The results showed that undersampled PAM data could be accurately restored, and the resolution was enhanced. Subsequently, we further verified the INR model using images of leaf veins and microvasculature of a mouse brain. The findings demonstrated that INR not only achieved high-fidelity restoration of feature sizes in undersampled AR-PAM data but also maintained good continuity in the images. These discoveries indicate that our method can significantly increase the speed of AR-PAM scanning while improving resolution, presenting broad prospects for application in biomedical imaging. The contributions of this paper are summarized as follows:
\begin{enumerate}
    \item For the first time, we consider the use of INR to simultaneously perform sparse image reconstruction and deconvolution for AR-PAM. This novel approach aims to enhance both the imaging speed and resolution of AR-PAM.

    \item In our method, the PSF of AR-PAM is learnable, which reduces the difficulty of accurately measuring the system's PSF. This feature makes our approach more flexible and adaptable, allowing it to automatically adjust and optimize the PSF under different imaging conditions. Consequently, this further enhances the quality and robustness of the image reconstruction.
\end{enumerate}

\section{Methods}

\subsection{AR-PAM Image Formation Process}
According to \cite{21}, the AR-PAM system can be modeled as a linear shift-invariant system around the focal region. The imaging quality of AR-PAM is governed by the principles of ultrasonic beamforming, which include the imaging characteristics such as lateral and axial resolution, as well as the depth of field. Therefore, the degradation model of the AR-PAM system can be derived. In practical beamforming scenarios, the field pattern of a plane circular transducer can be derived using the Rayleigh-Sommerfeld diffraction formula. This formula allows for the calculation of the field at any spatial point $\overrightarrow{r_t}$ as\cite{Lu1994BiomedicalUB,32}:

\begin{equation}
    \widetilde{\mathrm{E}}(\overrightarrow{r_{t}},\omega)=\frac{1}{i\lambda}\int_{0}^{R}\int_{-\pi}^{\pi}\widetilde{\epsilon}(r_{1},\omega)e^{i\omega r_{01}/c}\frac{z}{r_{01}^{2}}r_{1}d\varphi_{1}dr_{1}
\end{equation}

\begin{equation}
    r_{01}=\sqrt{(x_t-r_1\cos\varphi_1)^2+(y_t-r_1\sin\varphi_1)^2+z^2}
\end{equation}

where $\lambda$ represents the wavelength, $\omega$ is the angular frequency, and $c$ denotes the speed of sound. The term $\omega / c$ corresponds to the wavenumber ($2\pi / \lambda$), and $r_{01}$ is the distance from the source point to the observation point in space. In this work, $\widetilde{\epsilon}(r_{1},\omega)$ refers to the Fourier transform of the aperture weighting function $(r_1, t)$, and $\widetilde{\mathrm{E}}(\overrightarrow{r_t},\omega)$ is the Fourier transform of the wavefield $\widetilde{\mathrm{E}}(\overrightarrow{r_t},\omega)$ at the observation point.

The field pattern of a spherically focused transducer can be obtained by combining the field pattern of a flat circular transducer with a spherical compensation function. This spherical compensation function acts as a spherical phase shifter along the radial distance and is specified as:

\begin{equation}
    \tilde{\epsilon}\left(r_{1},\omega\right)=e^{-i\omega(\sqrt{L^{2}+r_{1}^{2}}-L)/c}
\end{equation}

Assuming the spatial point is located on the X-axis, the double integral in (2) can be analytically derived as:

\begin{equation}
    \tilde{\mathrm{E}}\left(\vec{r}_{t},\omega\right)=\frac{\omega a^{2}}{icL}e^{i\omega\left(L+\frac{x_{t}^{2}}{2L}\right)}\left[\frac{2J_{1}\left(\omega x_{t}a/c\right)/L}{\left(\omega x_{t}a\right)/\left(cL\right)}\right]
\end{equation}

Here, $J_1$ is the first-order Bessel function of the first kind. The PSF (Point Spread Function) kernel $\kappa$ can be calculated as the integral of the field distribution of the spherically focused transducer, weighted by the transducer's spectrum $T(\omega)$, over the entire ultrasonic transducer bandwidth.

\begin{equation}
    \kappa=\int T(\omega)\tilde{\mathbb{E}}(\vec{r_{t}},\omega)d\omega
\end{equation}

In this work, the system's output can be described by its PSF. Specifically, if $f(x)$ represents the raw image acquired by the AR-PAM system and $g(x)$ denotes the true undistorted image, the acquisition process can be expressed by the following equation:

\begin{equation}
    g(x)=\kappa\otimes f(x)+n
\end{equation}

Here, $n$ indicates noise. The symbol $\otimes$ denotes the convolution operation. Since the focal region of the acoustic transducer used in AR-PAM typically exhibits a Gaussian profile, we assume that the PSF also follows a Gaussian distribution.

\subsection{low-sampling images}
The above operation can be considered as data obtained under dense full-scanning conditions. To speed up the acquisition of AR-PAM data, we can perform down-sampling to obtain sparsely sampled data\cite{29}. This process is illustrated in Figure 2. Specifically, using stride scaling, only half of the pixels along one lateral dimension  are selected and used in the down-sampled image. This means that the down-sampled image (in terms of pixels) contains only one-quarter of the pixels of the fully scanned image. Consequently, the image acquisition time for the first approximation can be theoretically reduced to one-fourth. Similarly, when a four-times stride scaling is applied, the down-sampled image contains only one-sixteenth of the pixels compared to the full scan situation, thus the image acquisition time is expected to be reduced to one-sixteenth of the full scan scenario. The down-sampling method used here will generate a good approximation of the down-sampled images obtained experimentally.

\begin{figure}[!t]
\centering
\includegraphics[width=2.3in]{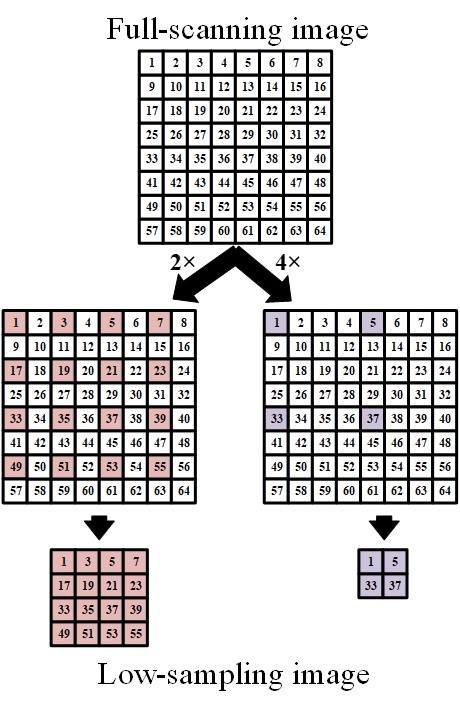}
\caption{The process of generating low-sampling images from the full-scanning ones.}
\label{fig_1}
\end{figure}

\subsection{Proposed Framework}
Our framework consists of four main components, as described in Figure 3. Firstly, we predict the high resolution initial acoustic pressure for dense sampling in AR-PAM at given spatial positions $\mathbf{x} \in \mathbb{R}^2$. Secondly, we convolve the predicted image with the PSF to obtain a prediction of the image that would be acquired under dense sampling conditions in real AR-PAM. Thirdly, we uniformly down-sample the dense sampling image to simulate an image acquired under sparse sampling conditions in real AR-PAM. Finally, backpropagation of the computed loss function is used to update the parameters within the INR.

\begin{figure*}[!t]
    \centering
    \includegraphics[width=7.4in]{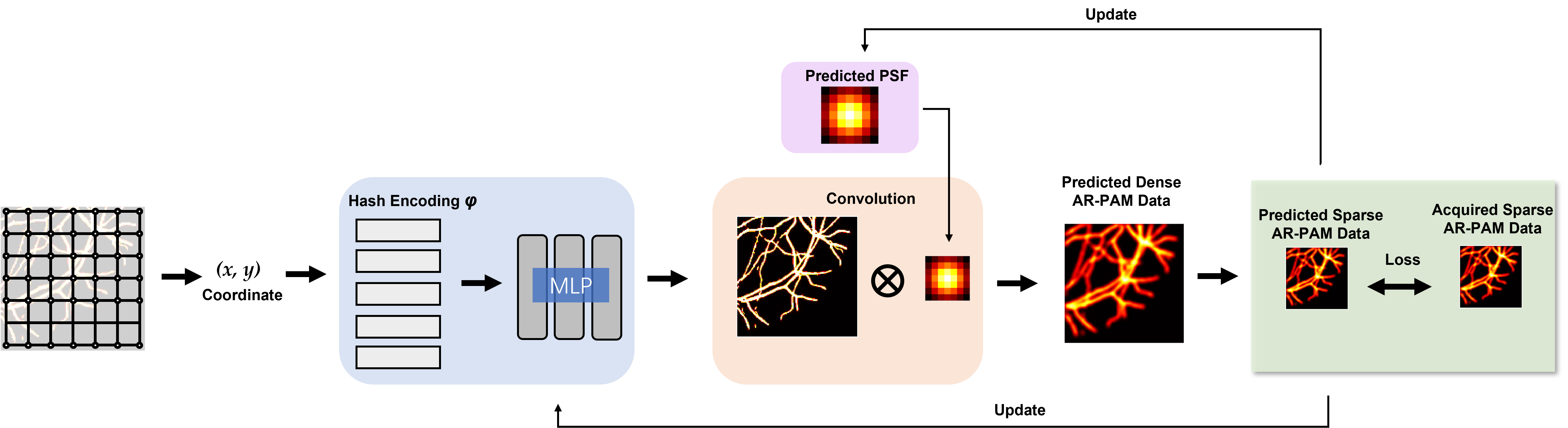}%
    \label{fig:example}
    \centering
    \caption{The pipeline of proposed the sparse deconvolution reconstruction method. The network takes as input the coordinates $p$ of densely sampled high-resolution images and predicts the image intensities $I$ at these locations. Subsequently, PSF convolution and down-sampling operations are performed. Finally, the network is optimized by minimizing the loss between the predicted sparsely sampled AR-PAM images and the actual acquired AR-PAM images.}
    \end{figure*}

\begin{enumerate}    
    \item \textit{Hash Encoding:} In the first step, the input space is divided into an $m\times m$ grid, with each grid cell assigned a two-dimensional coordinate $(x, y)$. However, due to the spectral bias issue \cite{Rahaman2018OnTS}, fitting high-frequency signals using only two-dimensional coordinates is extremely challenging. To mitigate this problem, various encoding strategies \cite{Tancik2020FourierFL,Sun2021CoILCI,33} have been proposed to map low-dimensional inputs to high-dimensional feature vectors, which enables subsequent MLP to easily capture high-frequency components and thereby reduce approximation errors. In our method, we employ hash encoding. Hash encoding assigns a trainable feature to each input coordinate, transforming low-dimensional coordinates into high-dimensional features. This adaptive encoding strategy is task-specific and enables powerful fitting capabilities even with a relatively shallow MLP. 
    
    \item \textit{Three-Layers MLP:} After hash encoding, the two-dimensional input coordinate $p \in \mathbb{R}^2$ is encoded into a high-dimensional feature vector. Subsequently, a 3-layer MLP is used to transform the feature vector into image intensity $I$. The MLP consists of two hidden layers, each with 64 neurons and using the ReLU activation function, and an output layer with a Sigmoid activation function. Based on this encoding, the MLP predicts the initial acoustic pressure values with high resolution for dense sampling in AR-PAM. 
    \item \textit{Learnable PSF:} The initial acoustic pressure image is then convolved with the learnable PSF to generate data that simulates real AR-PAM acquisition conditions. The PSF is initially estimated based on the measurements from the AR-PAM system. This initial PSF serves as a starting point. Then, in each iteration, the PSF is refined through a learning process.
\end{enumerate}

To simulate the undersampling imaging process, we uniformly down-sample the densely sampled image. To constrain the INR, we include two different loss functions.

\begin{equation}
    \mathcal{L}=\sum_{i\in I}\left(I'(i)-I(i)\right)^2+\epsilon TV(S)
\end{equation}

We utilize the L2 loss to measure the discrepancy between the predicted image \( I' \) and the acquired corresponding image \( I \). Additionally, we use total variation loss to regularize the predicted high-resolution initial acoustic pressure map of dense sampling in AR-PAM. This approach has been proven effective in the regularization of deconvolution algorithms, and it provides us with improved results\cite{30}.

\subsection{Blind Deconvolution algorithm}
As a comparison, we use the blind deconvolution algorithm. Currently, most blind deconvolution methods fall under the variational Bayesian inference framework \cite{Chowdhury2020NonblindAB}. The primary differences among these methods lie in the form of the likelihood function, the choice of prior for the target, and the methods used to estimate the blur kernel and find the optimal solution. Here, we adopt a general blind deconvolution method, as referenced in \cite{Chowdhury2020NonblindAB}, which employs expectation-maximization optimization to find the maximum posteriori solution under a flat prior. Blind deconvolution is performed on each input blurred image using MATLAB, with 30 iterations.

\subsection{Data}
\textbf{In simulated data:} To compare our method with the baseline, we utilized simulated vascular data. To demonstrate the effectiveness of our approach, we compared the results with various interpolation deconvolution methods.

\textbf{In leaf vein data:} To further validate the feasibility of our method, we used experimentally acquired sparse PAM images to make the scenario closer to practical applications. In this dataset\cite{29}, AR-PAM scanned the same region of interest (ROI) with scanning strides of 8$\mu\mathrm{m}$ and 16$\mu\mathrm{m}$. In this demonstration, we obtained 256$\times$256 pixel fully sampled PAM images and corresponding 128$\times$128 (or 64$\times$64 pixel) low-sampled PAM images for the same ROI. The low-sampled PAM images were used as input, while the 256$\times$256 pixel images served as references.

\textbf{In vivo mouse brain microvasculature data:} We utilized \textit{in vivo} mouse brain microvascular system data obtained by the Duke University Photoacoustic Imaging Laboratory, which were acquired using the PAM system previously published in \cite{31}. The PAM system has a lateral resolution of 5$\mu\mathrm{m}$ and an axial resolution of 15$\mu\mathrm{m}$. Since the data were acquired using OR-PAM, according to the study by Zhang et al. \cite{32}, the data can be degraded from OR-PAM to AR-PAM resolution. Therefore, for this dataset, we consider the data obtained by OR-PAM as the ground truth.

\begin{figure*}[b]
    \centering
    \includegraphics[width=6.5in]{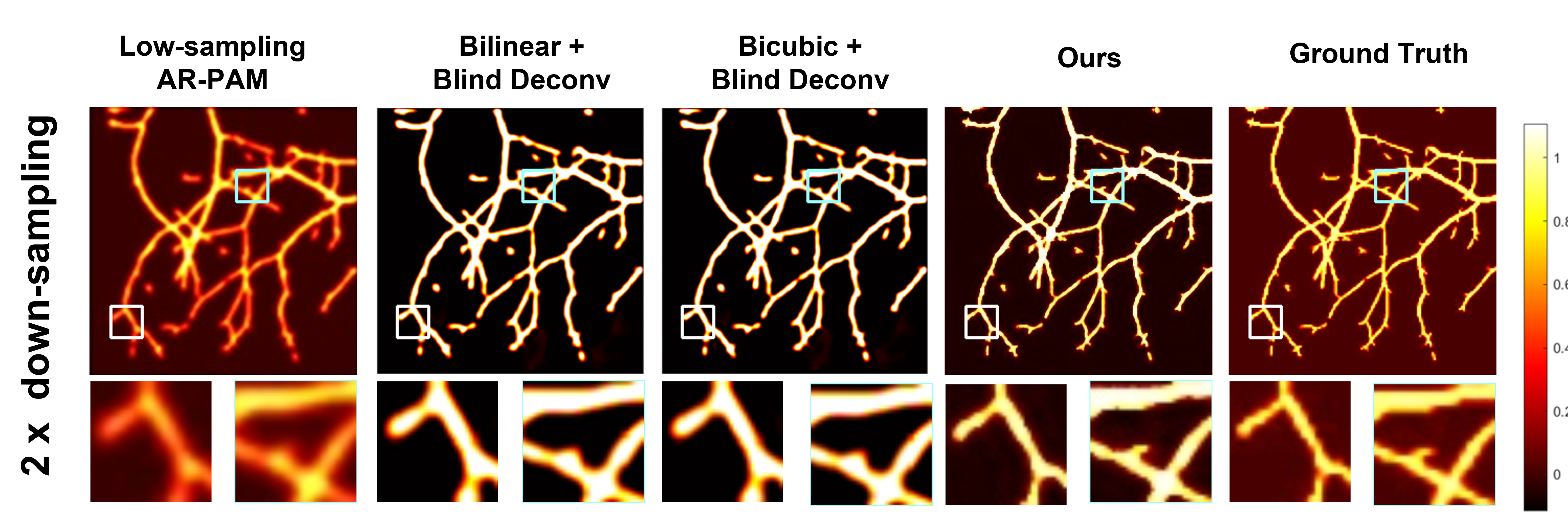}%
    \label{fig:example}
    \centering
    \caption{Qualitatively compare the deblurring and sparse reconstruction performance of our method with that of blind deconvolution using different interpolation techniques. Examples of simulated vascular images at 2x down-sampling are provided.}
\end{figure*}

\begin{figure*}[b]
    \centering
    \includegraphics[width=6.5in]{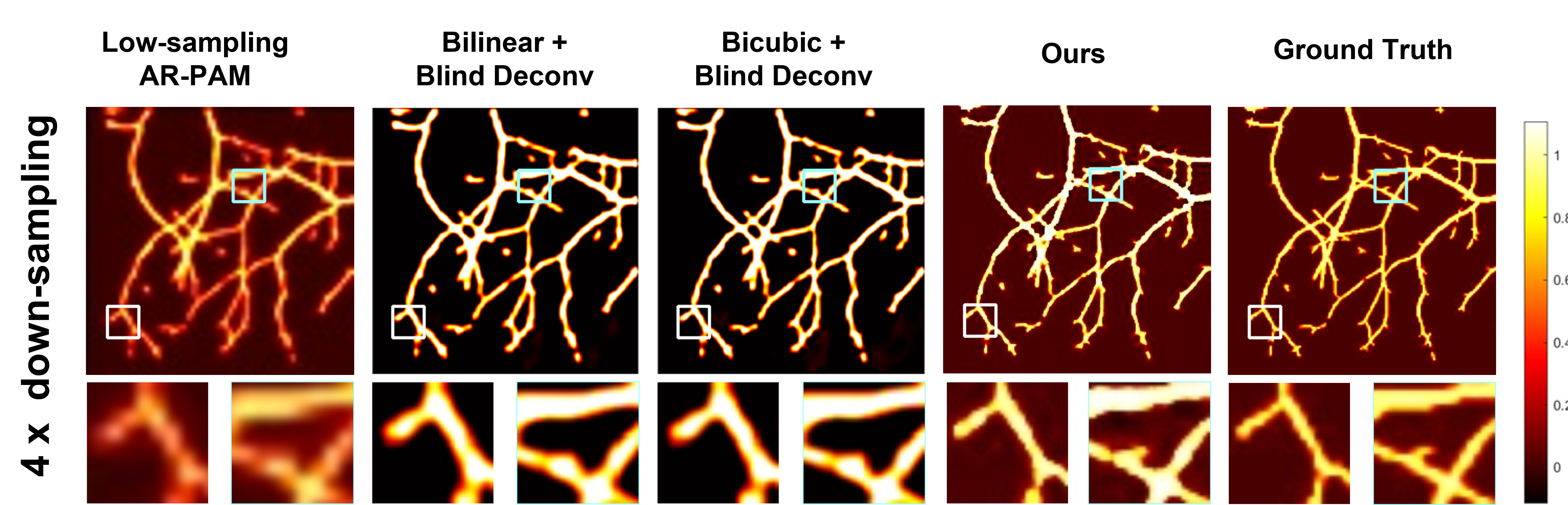}%
    \label{fig:example}
    \centering
    \caption{Qualitatively compare the deblurring and sparse reconstruction performance of our method with that of blind deconvolution using different interpolation techniques. Examples of simulated vascular images at 4x down-sampling are provided.}
\end{figure*}

\section{Results}
\subsection{Implementation Details}
The proposed framework is entirely implemented in PyTorch. Using hash encoding\cite{33}, the MLP can remain relatively small; therefore, we set it to have 3 layers with 64 neurons per layer. We employed the Adam optimizer \cite{34} to minimize the loss function, with hyperparameters set as follows: $\beta_1 = 0.9$, $\beta_2 = 0.999$, and $\epsilon = 10^{-8}$. The initial learning rate was set to $10^{-3}$, with a decay factor of 0.5 every 500 epochs. The total number of training epochs was 5000, and the entire training process took approximately 30 seconds on a single NVIDIA 1080 Ti GPU. Notably, all the training parameters remained the same for different input views.

\subsection{Stimulation data results}

In Figures 4 and 5, we compare the GT with the predictions from different interpolation deconvolution methods and our proposed method. From these figures, it can be qualitatively observed that our proposed method generates higher-resolution images, providing sharper edges and achieving clearer target separation. It is evident that at a 4x down-sampling rate, traditional interpolation methods cannot adequately recover the data of unmeasured points, leading to subsequent deconvolution operations failing to effectively enhance resolution, resulting in blurry edges. 

Through observation, it can be seen that our proposed method retains image details while better reconstructing the true shape and boundaries of the target objects, making the images visually clearer and more recognizable. In contrast, traditional interpolation deconvolution methods may lead to blurriness or unclear boundaries between targets, which could result in misinterpretation or inaccurate interpretation of the image content in certain situations.

Table 1 quantitatively measures the reconstruction effects of various methods. Compared to two traditional interpolation-based deconvolution methods, our method achieves an increase of 8.81 dB and 8.77 dB in PSNR under 2x down-sampling, while SSIM improves by 0.0847 and 0.0774, respectively. As the down-sampling rate increases, the advantages of our method become more pronounced. At 4x down-sampling, the PSNR increases by 6.52 dB and 6.78 dB, while SSIM improves by 0.1921 and 0.1916, respectively.

%经典三线表
\begin{table}[H]
\caption{\textbf{Quantitative Comparison of Different Methods for Simulated Vessel Sparse Deconvolution Reconstruction Using SSIM and PSNR}}
\centering
\begin{tabular}{ccccc}%四个c代表有四列且内容居中
\toprule%第一道横线
&\multicolumn{1}{c}{\textbf{\underline{2X}}}& &{\textbf{\underline{4X}}} \\%跨两列;内容居中;跨列内容为Resultsummary通过\textbf{}与\underline{}命令分别对内容加粗、加下划线
Case&PSNR&SSIM&PSNR&SSIM \\

\midrule%第三道横线 
Bilinear+Blind Deconv&21.23&0.8701&16.96&0.7207 \\

\midrule%第三道横线 
Bicubic+Blind Deconv&21.27&0.8774&16.70&0.7212 \\

\midrule%第三道横线 
Ours&30.04&0.9548&23.48&0.9128 \\

\bottomrule%第四道横线
\end{tabular}
\end{table}

\subsection{Leaf vein experimental results}
Subsequently, to further validate our findings, we performed validation experiments using a publicly available OR-PAM leaf vein dataset. In these experiments, we utilized the PSF of the AR-PAM system to generate low-resolution images, with OR-PAM serving as the ground truth. To explore the robustness and adaptability of our method in practical AR-PAM scenarios, we implemented two different down-sampling rates, reflecting the varying levels of image resolution degradation.

Our results, presented in Figures 6 and 7, qualitatively demonstrate that even under challenging conditions such as reduced resolution, our method can clearly depict the leaf vein structures, including their fine contours. Notably, the internal details of the veins, which are often lost or severely degraded in traditional blind deconvolution methods, are preserved and restored using our proposed method. We also quantitatively compared the recovery performance of different methods, and the results are summarized in Table II. At 2x down-sampling, our method achieves PSNR improvements of 2.71 dB and 2.59 dB compared to the two traditional methods. Additionally, SSIM increases by 0.2043 and 0.1954, respectively. At 4x down-sampling, the PSNR improves by 4.45 dB and 4.41 dB, while SSIM increases by 0.5009 and 0.4971, respectively. These comparisons reveal that our method not only achieves clearer distinction between adjacent veins but also significantly sharpens the edges, thereby enhancing the resolution and overall structural clarity.

%经典三线表
\begin{table}[H]
    \caption{\textbf{Quantitative Comparison of Different Methods for Leaf Vein Sparse Deconvolution Reconstruction Using SSIM and PSNR}}
    \centering
    \begin{tabular}{ccccc}%四个c代表有四列且内容居中
    \toprule%第一道横线
    &\multicolumn{1}{c}{\textbf{\underline{2X}}}& &{\textbf{\underline{4X}}} \\%跨两列;内容居中;跨列内容为Resultsummary通过\textbf{}与\underline{}命令分别对内容加粗、加下划线
    Case&PSNR&SSIM&PSNR&SSIM \\

    \midrule%第三道横线 
    Bilinear+Blind Deconv&16.30&0.5623&13.65&0.2209 \\
    
    \midrule%第三道横线 
    Bicubic+Blind Deconv&16.42&0.5712&13.69&0.2247 \\
    
    \midrule%第三道横线 
    Ours&19.01&0.7666&18.10&0.7218 \\
    
    \bottomrule%第四道横线
    \end{tabular}
\end{table}

\begin{figure}[H]
\centering
\includegraphics[width=2.7in]{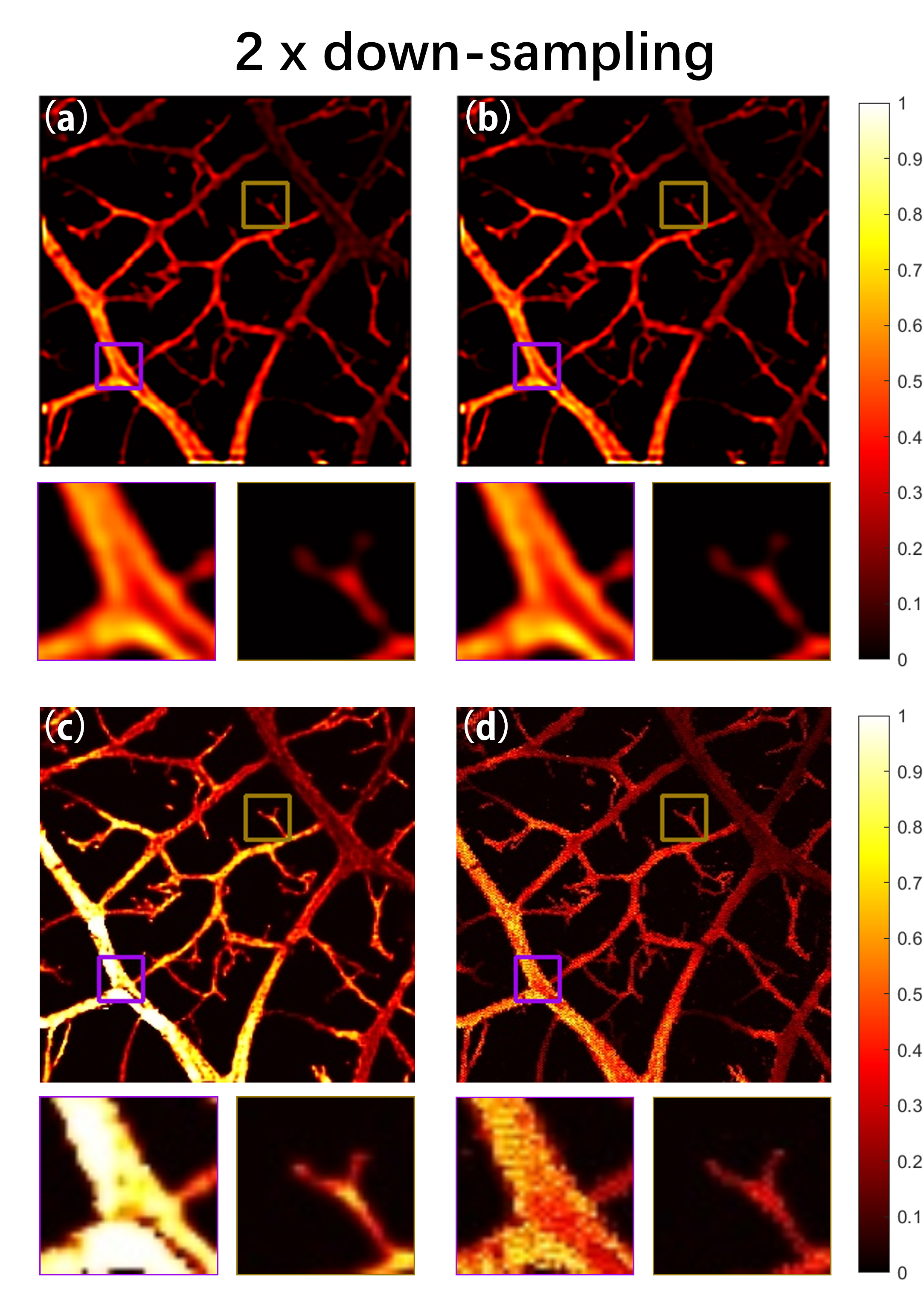}
\caption{Qualitatively compare the deblurring and sparse reconstruction performance of our method with that of blind deconvolution using different interpolation techniques. Examples of leaf vein data at 2x down-sampling. (a) by bilinear+blind deconvolution, (b) by bicubic+blind deconvolution, (c) by our method, (d) OR-PAM(Ground truth).}
\label{fig_1}
\end{figure}

\begin{figure}[H]
\centering
\includegraphics[width=2.7in]{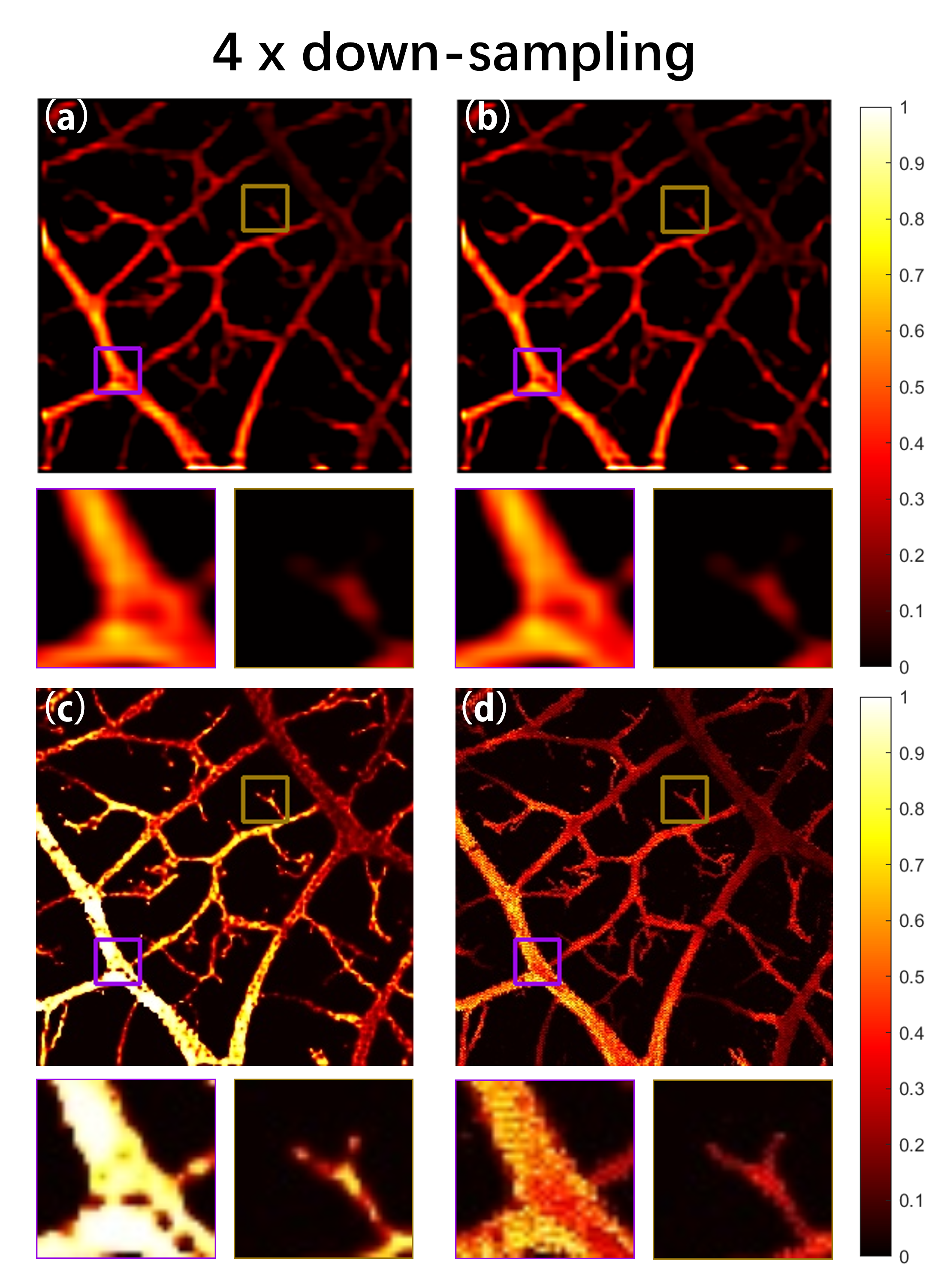}
\caption{Qualitatively compare the deblurring and sparse reconstruction performance of our method with that of blind deconvolution using different interpolation
techniques. Examples of leaf vein data at 4x down-sampling. (a) by bilinear+blind deconvolution, (b) by bicubic+blind deconvolution, (c) by our method, (d) OR-PAM(Ground truth).}
\label{fig_1}
\end{figure}

% \begin{figure*}[!t]
% \centering
% \includegraphics[width=5.0in]{fig6.jpg}%
% \label{fig:example}
% \centering
% \caption{(a) Qualitative results of deconvolution for all comparative methods on leaf vein  Data at a 2x downsampling. (b) 1D profiles along the white lines.}  
% \end{figure*}

% \begin{figure*}[!t]
% \centering
% \includegraphics[width=5.0in]{fig8.png}%
% \label{fig:example}
% \centering
% \caption{(a) Qualitative results of deconvolution for all comparative methods on leaf vein  Data at a 4x downsampling. (b) 1D profiles along the white lines.}  
% \end{figure*}

\subsection{In vivo experiment}

To further validate the reconstruction ability of our method, we applied it to publicly available \textit{in vivo} mouse brain vasculature images. In this set of experiments, the mouse brain vasculature data were acquired using OR-PAM, thus having high resolution, which can be considered as the ground truth. According to Ref.\cite{32}, we can degrade OR-PAM to AR-PAM, and since the degradation process remains a convolution, our method still applies. Figure 8 shows the results of different methods and the high-resolution images from OR-PAM. Green boxes indicate the ROI, which we compared. It can be observed that our proposed method yields sharper vessel patterns and enhanced vessel signals, with clearer edges. In addition to qualitative analysis, we performed quantitative analysis, and Table III presents the PSNR and SSIM results. At 2x down-sampling, our method achieves PSNR improvements of 6.04 dB and 6.12 dB compared to the two traditional methods, with SSIM increases of 0.1694 and 0.1697, respectively. At 4x down-sampling, the PSNR improves by 5.1 dB and 5.44 dB, while SSIM increases by 0.1795 and 0.1761, respectively.

%经典三线表
\begin{table}[H]
    \caption{\textbf{Quantitative Comparison of Different Methods for Mouse Brain Vasculature Sparse Deconvolution Reconstruction Using SSIM and PSNR}}
    \centering
    \begin{tabular}{ccccc}%四个c代表有四列且内容居中
    \toprule%第一道横线
    &\multicolumn{1}{c}{\textbf{\underline{2X}}}& &{\textbf{\underline{4X}}} \\%跨两列;内容居中;跨列内容为Resultsummary通过\textbf{}与\underline{}命令分别对内容加粗、加下划线

    Case&PSNR&SSIM&PSNR&SSIM \\

    \midrule%第三道横线 
    Bilinear+Blind Deconv&19.50&0.7335&20.22&0.6803 \\
    
    \midrule%第三道横线 
    Bicubic+Blind Deconv&19.42&0.7332&19.88&0.6837 \\
    
    \midrule%第三道横线 
    Ours&24.54&0.9029&25.32&0.8598 \\
    
    \bottomrule%第四道横线
    \end{tabular}
\end{table}

\begin{figure}[h]
\centering
\includegraphics[width=3.5in]{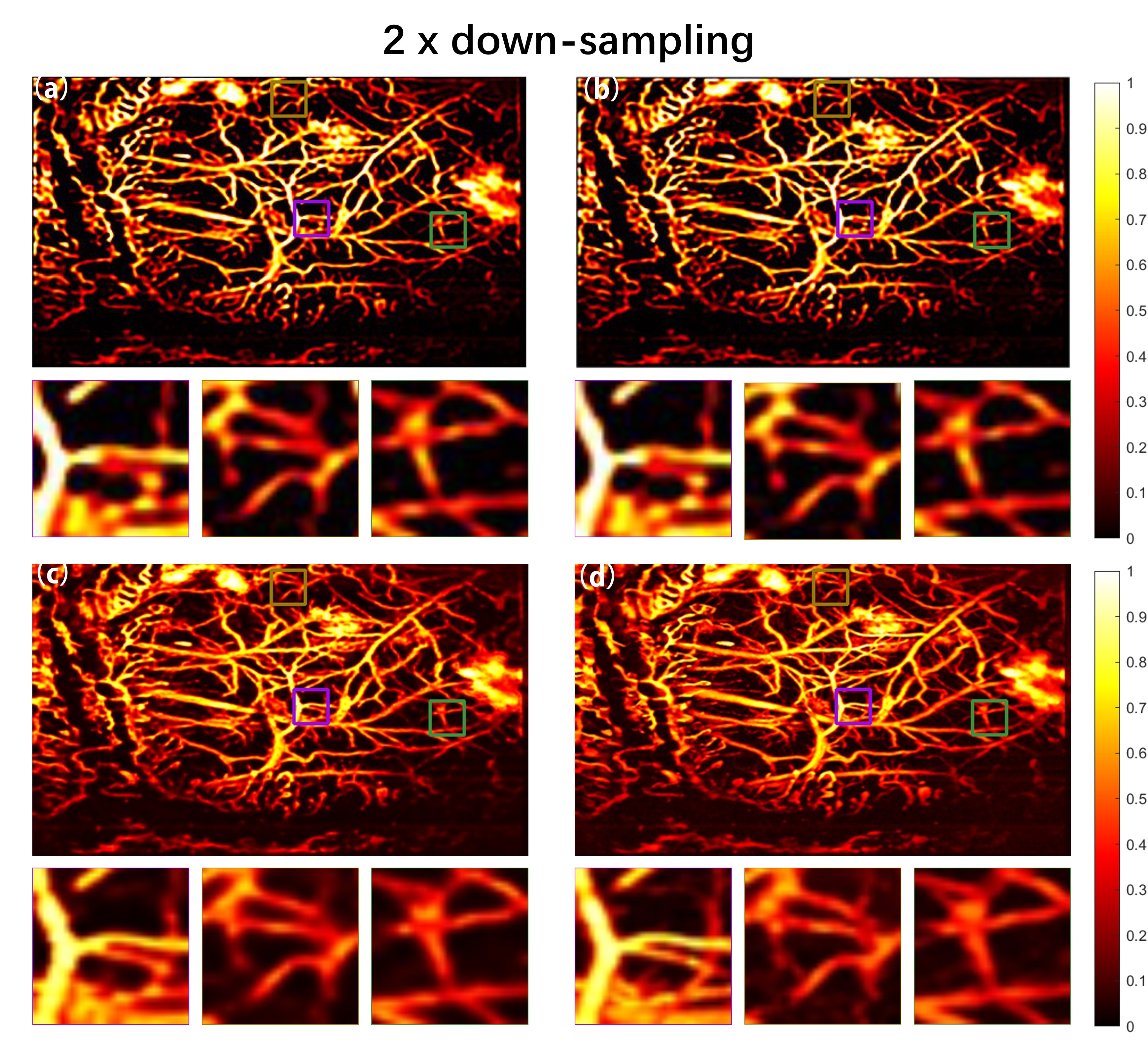}
\caption{Qualitatively compare the deblurring and sparse reconstruction performance of our method with that of blind deconvolution using different interpolation
techniques. Examples of \textit{in vivo} mouse brain microvasculature data at 2x down-sampling. (a) by bilinear+blind deconvolution, (b) by bicubic+blind deconvolution, (c) by our method, (d) OR-PAM(Ground truth).}
\label{fig_1}
\end{figure}
    
\begin{figure}[h]
\centering
\includegraphics[width=3.5in]{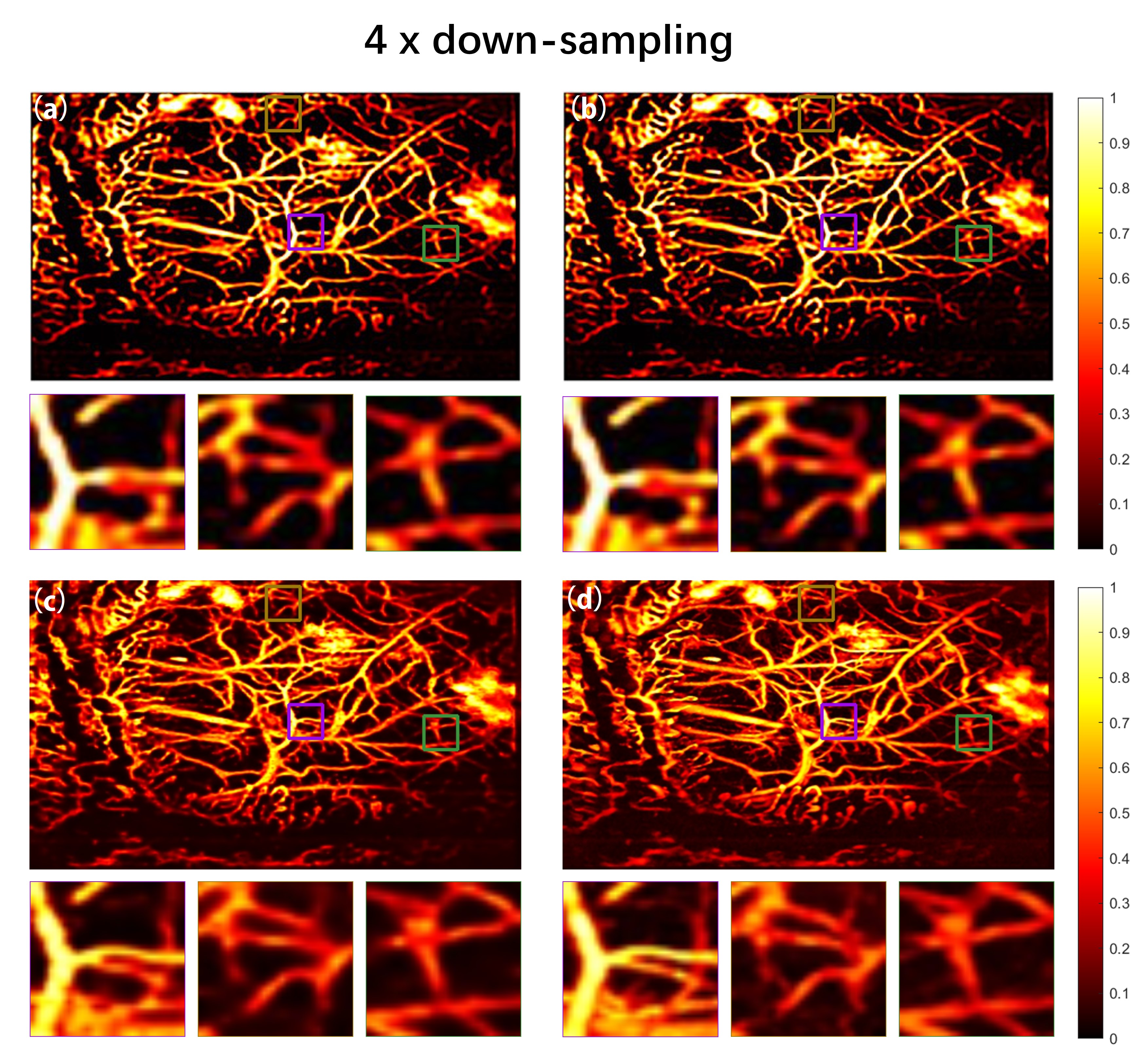}
\caption{Qualitatively compare the deblurring and sparse reconstruction performance of our method with that of blind deconvolution using different interpolation
techniques. Examples of \textit{in vivo} mouse brain microvasculature data at 4x down-sampling. (a) by bilinear+blind deconvolution, (b) by bicubic+blind deconvolution, (c) by our method, (d) OR-PAM(Ground truth).}
\label{fig_1}
\end{figure}

\section{Discussion}
In this work, we investigate sparse deconvolution techniques based on INR to enhance the resolution of AR-PAM images. The novelty of this work lies in the first application of INR methods to improve the resolution of focal region AR-PAM images. Compared to other deep learning methods, our approach does not require a large amount of paired data, thereby enhancing cost-effectiveness and reducing the difficulty of data acquisition. We evaluate our method against traditional interpolation and blind deconvolution methods. The results show that our method can achieve superior sparse deconvolution reconstruction, with sharper and clearer edges and more pronounced details compared to traditional methods.

The sparse deconvolution achieved through the INR method can be explained as follows: First, each pixel in the high-resolution densely sampled image is assigned a two-dimensional coordinate, which is then hashed and passed as input to the INR network. Our network consists of two hidden layers, each containing 64 parameters. Finally, the network outputs the predicted initial acoustic pressure at each location. After obtaining the predicted initial acoustic pressures for all locations, convolution and down-sampling operations are performed, and loss calculation and backpropagation are conducted based on the actual acquired data. Notably, since it is difficult to obtain an exact PSF in practical operations, our network incorporates a learnable PSF, allowing for better PSF estimation. Additionally, due to the use of hashing, the entire reconstruction process is very fast, taking only 30 seconds over 5,000 iterations. Compared to supervised deep learning methods that require a large amount of paired training data, our method demonstrates better generalization, as supervised networks may perform poorly on unseen data types.

Given that AR-PAM is often used for vascular imaging, we chose simulated vascular data for our experiments. The results show that, compared to other non-data-driven methods, our method provides clearer sparse reconstruction and deconvolution, reducing image blurring. To further validate the effectiveness of our method, we also used real leaf vein and mouse cerebral vasculature data. The results demonstrate that our method maintains excellent reconstruction capabilities, significantly enhancing image details and resolution. 

\section{Conclusion}
In this work, we investigated sparse deconvolution techniques based on INR and proposed a deep learning framework that leverages INRs. We compared our method with traditional interpolation-based blind deconvolution methods through both simulation and \textit{in vivo} experiments. Compared to all interpolation-based blind deconvolution methods, our approach provides superior resolution and image detail. When applied to \textit{in vivo} images of mouse brain vasculature, our method achieved high-fidelity deconvolution of sparse vessels while maintaining good continuity. This work holds promise for enhancing the resolution of AR-PAM images while accelerating data acquisition. The proposed method can be generalized and applied to other imaging modalities to enhance their resolution.

\bibliographystyle{unsrt}  % 使用 BibTeX 样式
\bibliography{ref}        % 你的 .bib 文件名，不需要 .bib 扩展名

\end{document}